\documentclass[journal,transmag]{IEEEtran}
\usepackage{booktabs}
\usepackage{tabularx}

\usepackage{graphicx}

\usepackage{caption,subcaption}
\usepackage{amsmath, amssymb, latexsym}
\DeclareMathOperator{\ReLU}{ReLU}
\newcommand{\KL}{\ensuremath{\mathrm{KL}}}

\usepackage{tikz}
\usepackage{xcolor}
\usepackage{lipsum}
\usepackage{float}
\definecolor{fc}{HTML}{8bd0ff}
\definecolor{h}{HTML}{228B22}
\definecolor{bias}{HTML}{87CEFA}
\definecolor{noise}{HTML}{8B008B}
\definecolor{conv}{HTML}{eeeff3}
\definecolor{pool}{HTML}{fdf6f4}
\definecolor{up}{HTML}{fdf6f4}
\definecolor{view}{HTML}{FFFFFF}
\definecolor{bn}{HTML}{FFD700}
\tikzset{fc/.style={black,draw=black,fill=fc,rectangle,minimum height=1cm}}
\tikzset{h/.style={black,draw=black,fill=h,rectangle,minimum height=1cm}}
\tikzset{bias/.style={black,draw=black,fill=bias,rectangle,minimum height=1cm}}
\tikzset{noise/.style={black,draw=black,fill=noise,rectangle,minimum height=1cm}}
\tikzset{conv/.style={black,draw=black,fill=conv,rectangle,minimum height=1cm}}
\tikzset{pool/.style={black,draw=black,fill=pool,rectangle,minimum height=1cm}}
\tikzset{up/.style={black,draw=black,fill=up,rectangle,minimum height=1cm}}
\tikzset{view/.style={black,draw=black,fill=view,rectangle,minimum height=1cm}}
\tikzset{bn/.style={black,draw=black,fill=bn,rectangle,minimum height=1cm}}
 
\usepackage{xspace}

\ifCLASSINFOpdf
\else
\fi
\begin{document}
%
\title{Soccer Event Detection Using Deep Learning}


\author{\IEEEauthorblockN{Ali Karimi\IEEEauthorrefmark{1},
Ramin Toosi\IEEEauthorrefmark{1},
Mohammad Ali Akhaee\IEEEauthorrefmark{1}}
\IEEEauthorblockA{\IEEEauthorrefmark{1}School of Electrical and Computer Engineering, College of Engineering, University of Tehran, Tehran, Iran.}

\thanks{
Corresponding author:Mohammad Ali Akhaee (email :akhaee@ut.ac.ir).}}


%



\IEEEtitleabstractindextext{%
\begin{abstract}
Event detection is an important step in extracting knowledge from the video. In this paper, we propose a deep learning approach to detect events in a soccer match emphasizing the distinction between images of red and yellow cards and the correct detection of the images of selected events from other images. This method includes the following three modules: i) the variational autoencoder (VAE) module to differentiate between soccer images and others image, ii) the image classification module to classify the images of events, and iii) the fine-grain image classification module to classify the images of red and yellow cards. Additionally, a new dataset was introduced for soccer images classification that is employed to train the networks mentioned in the paper. In the final section, 10 UEFA Champions League matches are used to evaluate the networks' performance and precision in detecting the events. The experiments demonstrate that the proposed method achieves better performance than state-of-the-art methods.
\end{abstract}

\begin{IEEEkeywords}
Event detection, Neural networks, Soccer video
\end{IEEEkeywords}}

\maketitle

\IEEEdisplaynontitleabstractindextext

%
\IEEEpeerreviewmaketitle

\section{Introduction}
%
%
%
%
\IEEEPARstart{S}{}occer is among the most popular sports in the world. The attractiveness of this sport has gathered many spectators. Various studies are being performed in this area to grow and assist this sport and meet the needs of soccer clubs and media. These researches mainly focus on estimating team tactics \cite{suzuki2019team}, tracking of players \cite{manafifard2017survey} or ball on the field \cite{kamble2019deep}, detection of events occurring in the match \cite{hong2018end,fakhar2019event,khan2018soccer,khan2018learning,jiang2016automatic}, summarizing the soccer match \cite{agyeman2019soccer,sanabria2019deep,rafiq2020scene} and estimating ball possession statistics \cite{sarkar2019generation}. These studies are carried out using various methods and techniques, including machine learning (ML).

Artificial intelligence (AI), and more specifically ML can assist in conducting the above-mentioned researches in order to achieve better and more intelligent results. ML itself can be implemented in a variety of ways, including deep learning (DL). Distinguishing the events of a soccer match is one of the active research fields related to soccer. Today, various AI methods are utilized to detect events in a soccer game \cite{hong2018end,khan2018learning}. The use of AI in this area can help to achieve higher accuracy in detecting events.

Detecting the events of a soccer match may have different applications. Event detection, for instance, in a soccer match can help obtaining the statistics of events of the match. Counting the number of free kicks, fouls, tackles, etc. in a soccer game can be done by a manpower. Using a manpower is not only costly and time consuming, but also is may be associated with errors. Nonetheless, with intelligent systems based on event detection, statistics can be calculated and used automatically. Other applications of event detection may be the summarization of a soccer match. \cite{zawbaa2012event}. The summary of a soccer match includes important events in which the match took place. To prepare a useful summary of a soccer match, the events should be correctly identified. Using a method for identifying the events with high accuracy can improve the quality of the summarization task.

The purpose of the current study is to detect events in soccer matches. In this regard, various deep learning architectures have been developed. When using deep learning methods, there always exists a need for datasets for training. To this end, first, a rich visual dataset is collected for our task such as (penalty kick, corner kick, free kick, tackle, to substitute, yellow and red cards) and images containing these events of the sides and center of the field are collected. Then, this image dataset is used to train the proposed networks. Solving event detection problem encounters some challenges including the similarity of some images such as yellow and red cards, which leads to some difficulties in separating the images of these two groups from each other. This causes the classifier to have trouble in distinguishing between yellow and red cards, resulting in incorrect detection.

Another problem in detecting events that usually occurs in a soccer match is the issue of no highlights. Not all events happening in a soccer match can be included in specific events to be given to the network for training. Some events that have not been fed to the network before training the network may occur specifically in a soccer match. In such cases, the network must be able to face these images or videos and not mistakenly categorize them in one of the given events. We solved the problem of \textit{no highlight} image detection by using VAE , setting a threshold and using additional classes in the image classification module.

The proposed method for detecting the soccer match events uses two convolutional neural network (CNN) and a variational autoencoder (VAE). The current study focuses on resolving the problem of no highlight frames which may be wrongly considered as one of the events, as well as dealing with the problem of similarity between red and yellow cards frames.  Experiments demonstrate the proposed method improves the accuracy of image classification from 88.93\% to 93.21\%. The main reason for increasing the accuracy is the use of a new fine-grain classification network to classify yellow and red cards. Also, the proposed method has a good ability to detect no highlight frames with a high precision. All datasets and implementations are publicly available.\footnote{https://github.com/FootballAnalysis/footballanalysis}

The rest of the paper is organized as follows.  Section II provides a literature review and examines the drawbacks of the works done in this area. Section III describes the proposed algorithm, the mechanism of its structure and working. Section IV introduces the datasets collected for this study. Section V presents the experimental results and compares the results with those of other papers. Eventually, Section VI concludes the paper.

\section{Related Work}

The research of Duan et. al. \cite{duan2005unified} is one of the first works in this area that implements supervised learning for top-down video shot classification. The authors in \cite{tavassolipour2013event} present a method for detecting soccer events using the Bayesian network. The basic methods presented suffered from low accuracy, until some methods have been proposed using DL. By presenting a method based on the convolutional network and the LSTM network, Agyeman et. al. \cite{agyeman2019soccer} present a method for summarizing a soccer match based on event detection, in which five events including corner kicks, free kicks, goal scenes, centerline, and throw-in are considered. This study uses 3D-ResNet34 architecture in the convolutional network structure. One of the problems with this work is that the number of events is limited and no highlights are taken into account. Jiang et. al. \cite{jiang2016automatic}, initially, perform feature extraction using a convolutional network, then perform event detection using its combination with RNN model.This method is limited to four events: goal, goal attempt, corner, and card. Sigari et. al. \cite{sigari2015fast} employ the fuzzy inference system. The algorithm presented in this method works based on replay detection, logo detection, view type recognition and audience excitement. This method is also limited to three events: penalty, corner, and free-kick. 11 events are classified in \cite{yu2019soccer}, covering a good number of events; while this method is not capable to distinguish between the red and yellow card events because of the high similarity between these two events. In general, the methods presented in this field work on either image, video \cite{agyeman2019soccer,jiang2016automatic}, or audio signals \cite{duxans2009audio,raventos2015automatic}. Nonetheless, there exist some methods employing two signals, i.e. audio and video, simultaneously \cite{sanabria2019deep,sigari2015fast}.

The methods recently presented utilize DL architectures as the main tool for feature extraction. Among the DL architectures suggested for feature extraction, the closest flagship architecture is EfficientNet architecture \cite{tan2019efficientnet}. This architecture is presented in 8 different versions. Different versions of this architecture offer generally higher performance than other previous models \cite{simonyan2014very,szegedy2015going,he2016deep}. Also, the proposed architecture has fewer parameters and occupies less memory.

One of the challenges of the event detection problem in soccer matches, which has not been addressed adequately in the literature, is the events that are very similar in appearance but are two separate events. For instance, in the images of yellow and red cards, only the color of the cards is different and the other parts of the image are the same. Although it may come to the mind that both are card-taking operations and may be almost the same, in the soccer game these two events impact the game process significantly. In the literature, both yellow and red card events are considered as one event \cite{tavassolipour2013event}, which causes problems for event detection. The reasons for this are the very high similarity of the images of these two events, which makes it very challenging to distinguish between the two events. In this paper, fine-grained image classification is used instead of common feature extraction architectures to solve this problem in detecting such events.

\begin{figure*}[h!]
\centering

\includegraphics[scale=0.4]{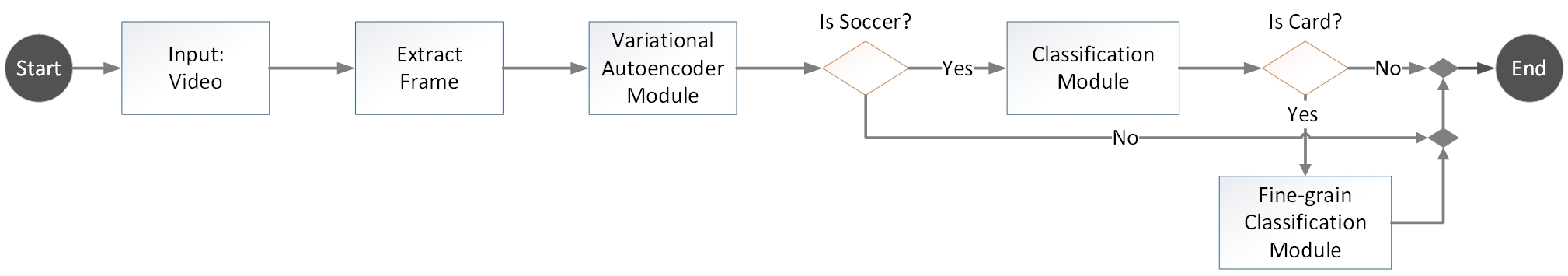}

\caption{Generic block diagram of the proposed algorithm}
\label{fig:1}
\end{figure*}

Fine-grained image classification is one of the challenges in machine vision, which categorizes images that fall into a similar category but do not fall into one subcategory \cite{dai2019bilinear}. For example, items such as face recognition, different breeds of dogs, birds, etc., despite the many structural similarities, do not fall into one subcategory and are different from each other. As another example, the California gull and Ringed-beak gull are two similar birds, differing only in beak pattern, and are in two separate subclasses. The main problem with this type of classification is that these differences are usually subtle and local. Finding these discriminating places of two subcategories is a challenge that we face in these methods.

The work of Lin et. al. \cite{lin2015bilinear} is one of the researches in the field of fine-grained image classification, which is based on deep learning. In this model, two neural networks are used simultaneously. The outer product of the outputs of these two networks is then mapped to a bilinear vector. Finally, there is a softmax layer to specify the classification of images. The accuracy of this method for the  CUB-200-2011 dataset \cite{wah2011caltech} is 84.1\% while using only the neural network architecture in this architecture provides the maximum accuracy of 74.7\%. Fu et. al. \cite{fu2017look} introduce a framework of recurrent attention CNN, which receives the image with the original size and passes it through a classification network, thereby extracts the probability of its placement in each category. At the same time, after the convolutional layers of the classifier, it extracts an attention proposal network that contains region parameters that it uses to zoom in on the image and then crop it. It now inputs the resulting new image like the original image into a network, and at the same time extracts an attention proposal network, thereby re-extracting another part of the new image. This method reaches an accuracy of 85.3\% \textit{}for the Birds dataset. In another work, the authors in \cite{zheng2017learning} propose a framework of multi-attention convolutional neural network with an accuracy of 86.5\%. Following in 2018, Sun et. al. \cite{sun2018multi} presented new architecture on an attention-based CNN that learns multiple attention region features per an image through the one-squeeze multi-excitation (OSME) module and then use the multi-attention multi-class constraint (MAMC). Thanks to this structure, this method improves the accuracy of previous methods to some extent. One of the latest method presented in \cite{ge2019weakly} consists of three steps. In the first step, the instances are detected and then instance segmentation is carried out. In the second step, a set of complementary parts is created from the original image. In the third step, CNN is used for each image obtained, and the output of these CNNs is given to the LSTM cells for all images. The accuracy of the best model in this method for the Birds dataset reaches 90.4\%. In general, the problem with the methods presented in this section is the accuracy they achieved, while newer methods attempt to improve the accuracy.

Another issue is that a soccer match can include various scenes that are not necessarily a specific event, such as scenes from a soccer match where players walking, or the moments when the game is stopped. Now, if such images are applied as input to the classification network, the network will mistakenly place them in one of the defined categories. The reason is that the network is trained only to categorize images between events, and is called traditional classification network \cite{geng2020recent}. In such classifications, the known class is used during training and the known class images should be given to the network during testing. Otherwise, the network will have trouble in detecting the image category, and even though the image should not be placed in any of the categories, it will be placed incorrectly in one of the defined categories. This type of categorization does not suffice for the problem under study. Because, as explained, the input images in this problem may not belong to any of the categories. Thus, we need a network to specify the category of an input image if it falls into one of the seven categories; otherwise, the network rejects it and does not mistakenly place it in these defined categories. In other words, an open set recognition is required to address the mentioned problem \cite{geng2020recent} .

Open set recognition techniques can be implemented using different methods. Cevikalp et. al. \cite{cevikalp2016best} performs this based on support vector machine (SVM). The works in \cite{hassen2020learning} and \cite{bendale2016towards} are also based on deep neural networks. Today, the use of generative models in this area is reaching its pinnacle, which is divided into two categories: instance generation and non-instance generation \cite{geng2020recent} . Finding a suitable method is still challenging, and the literature presented for event detection has not addressed this issue profoundly.

\section{Proposed method}

This section describes the proposed method. Initially, the general procedure is explained, then all three main parts of the method are introduced, which includes an image classification module for detecting the images of the defined events, a fine-grain classification module used to classify yellow and red card images, and a variational autoencoder for detecting \textit{no highlights}.

\subsection{The Proposed Algorithm}

As depicted in Fig \ref{fig:1}, the received video is first split into several frames based on the video length and frame rate per second, and then each frame is passed separately through a variational autoencoder. If loss value of the VAE network for the input frame is smaller than a specified value, the received frame is considered as an event frame and given to the image classification module. The image classification module classifies the images into nine classes. If the input image belongs to one of the center circle categories, right penalty area, or left penalty area, it is not categorized as an event and is still classified as no highlights. Yet, if it is one of the five events: penalty kick, corner kick, tackle, free kick, to substitute, it is recorded as an event in that image. If the event is a card event, the image is given to the fine-grain classification module to determine if it is a yellow card or red card, and the color of the card is specified there. Finally, to detect events in a soccer match, each event is calculated for every 15 consecutive frames (seven frames before, seven frames after, and the current frame), and if more than half of the frames belong to an event, that 15 frames (half a second) are tagged as that event. Of course, an event cannot be repeated more than once in 10 s, and if repeated, only one of them is calculated as an event in the calculation of the number of events occurred. In the folowing each module is described in detail.

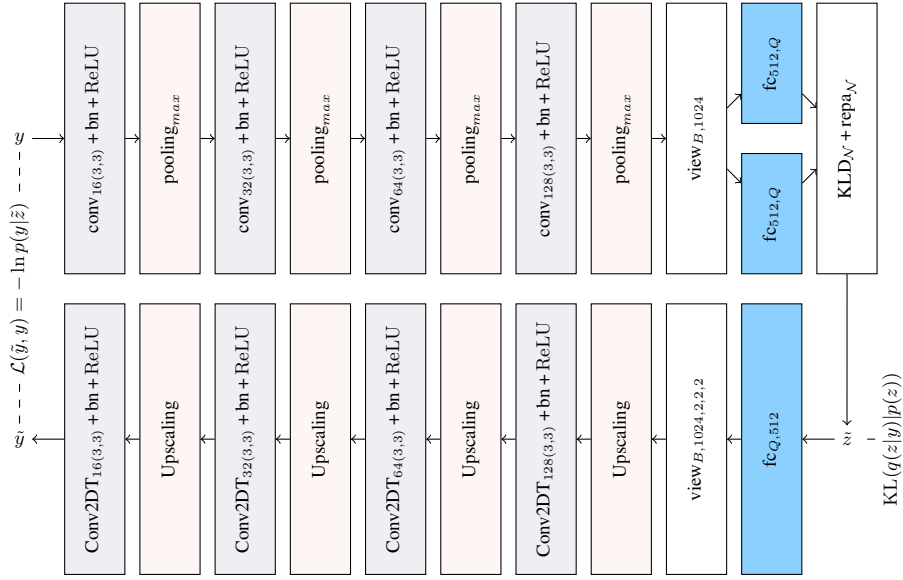
\begin{figure*}[hbt!]
  \centering
  \hspace*{-0.5cm}
  \begin{tikzpicture}[scale=0.8, transform shape]
    \node (y) at (0,0) {\small$y$};
 
    \node[conv,rotate=90,minimum width=4.5cm] (conv1) at (1.25,0) {\small$\text{conv}_{16(3,3) }$\,+\,$\text{bn}$\,+\,$\ReLU$};
    \node[pool,rotate=90,minimum width=4.5cm] (pool1) at (2.5,0) {\small$\text{pooling}_{max}$};
    \node[conv,rotate=90,minimum width=4.5cm] (conv2) at (3.75,0) {\small$\text{conv}_{32(3,3) }$\,+\,$\text{bn}$\,+\,$\ReLU$};
    \node[pool,rotate=90,minimum width=4.5cm] (pool2) at (5,0) {\small$\text{pooling}_{max}$};
    \node[conv,rotate=90,minimum width=4.5cm] (conv3) at (6.25,0) {\small$\text{conv}_{64(3,3) }$\,+\,$\text{bn}$\,+\,$\ReLU$};
    \node[pool,rotate=90,minimum width=4.5cm] (pool3) at (7.5,0) {\small$\text{pooling}_{max}$};
    \node[conv,rotate=90,minimum width=4.5cm] (conv4) at (8.75,0) {\small$\text{conv}_{128(3,3) }$\,+\,$\text{bn}$\,+\,$\ReLU$};
    \node[pool,rotate=90,minimum width=4.5cm] (pool4) at (10,0) {\small$\text{pooling}_{max}$};;
    
    \node[view,rotate=90,minimum width=4.5cm] (view4) at (11.25,0) {\small$\text{view}_{B, 1024}$};
    
    \node[fc,rotate=90, minimum width = 2cm] (fc51) at (12.5,1.25) {\small$\text{fc}_{512, Q}$};
    \node[fc,rotate=90, minimum width = 2cm] (fc52) at (12.5,-1.25) {\small$\text{fc}_{512, Q}$};
  
    \node[view,rotate=90,minimum width=4.5cm] (kld) at (13.75,0) {\small$\text{KLD}_{\mathcal{N}}$\,+\,$\text{repa}_{\mathcal{N}}$};
    
    \node (z) at (13.75,-5){\small$\tilde{z}$};
    
    \node[fc,rotate=90,minimum width=4.5cm] (fc6) at (12.5,-5) {\small$\text{fc}_{Q, 512}$};
    \node[view,rotate=90,minimum width=4.5cm] (view6) at (11.25,-5) {\small$\text{view}_{B, 1024, 2, 2, 2}$};
    
    \node[up,rotate=90,minimum width=4.5cm] (up7) at (10,-5) {\small$\text{Upscaling}$};
    \node[conv,rotate=90,minimum width=4.5cm] (conv7) at (8.75,-5) {\small$\text{Conv2DT}_{128(3,3)}$\,+\,$\text{bn}$\,+\,$\ReLU$};
    \node[up,rotate=90,minimum width=4.5cm] (up8) at (7.5,-5) {\small$\text{Upscaling}$};
    \node[conv,rotate=90,minimum width=4.5cm] (conv8) at (6.25,-5) {\small$\text{Conv2DT}_{64(3,3)}$\,+\,$\text{bn}$\,+\,$\ReLU$};
    \node[up,rotate=90,minimum width=4.5cm] (up9) at (5,-5) {\small$\text{Upscaling}$};
    \node[conv,rotate=90,minimum width=4.5cm] (conv9) at (3.75,-5) {\small$\text{Conv2DT}_{32(3,3)}$\,+\,$\text{bn}$\,+\,$\ReLU$};
    \node[up,rotate=90,minimum width=4.5cm] (up10) at (2.5,-5) {\small$\text{Upscaling}$};
    \node[conv,rotate=90,minimum width=4.5cm] (conv10) at (1.25,-5) {\small$\text{Conv2DT}_{16(3,3)}$\,+\,$\text{bn}$\,+\,$\ReLU$};
    
    \node (ry) at (0,-5) {\small$\tilde{y}$};
    
    \draw[->] (y) -- (conv1);
    \draw[->] (conv1) -- (pool1);
    \draw[->] (pool1) -- (conv2);
    
    \draw[->] (conv2) -- (pool2);
    \draw[->] (pool2) -- (conv3);
    
    \draw[->] (conv3) -- (pool3);
    \draw[->] (pool3) -- (conv4);
    
    \draw[->] (conv4) -- (pool4);
    \draw[->] (pool4) -- (view4);
    
    \draw[->] (view4) -- (fc51);
    \draw[->] (view4) -- (fc52);
    
    \draw[->] (fc51) -- (kld);
    \draw[->] (fc52) -- (kld);
    
    \draw[->] (kld) -- (z);
    \draw[->] (z) -- (fc6);
    \draw[->] (fc6) -- (view6);
    
    \draw[->] (view6) -- (up7);
    \draw[->] (up7) -- (conv7);
    
    \draw[->] (conv7) -- (up8);
    \draw[->] (up8) -- (conv8);
    
    \draw[->] (conv8) -- (up9);
    \draw[->] (up9) -- (conv9);
    
    \draw[->] (conv9) -- (up10);
    \draw[->] (up10) -- (conv10);
    
    \draw[->] (conv10) -- (ry);
 
    \node[rotate=90] (L) at (0, -2.5) {\small$\mathcal{L}(\tilde{y}, y) = -\ln p(y | \tilde{z})$};
    \draw[-,dashed] (y) -- (L);
    \draw[-,dashed] (ry) -- (L);
 
    \node[rotate=90] (KLD) at (14.5, -5) {\small$\KL(q(z|y) | p(z))$};
    \draw[-,dashed] (KLD) -- (z);
  \end{tikzpicture}
  \vskip 6px
  \caption{No highlight detection module architecture (VAE)}
  \label{fig:2}
\end{figure*}

\begin{figure*}[h!]
\centering

\includegraphics[scale=0.4]{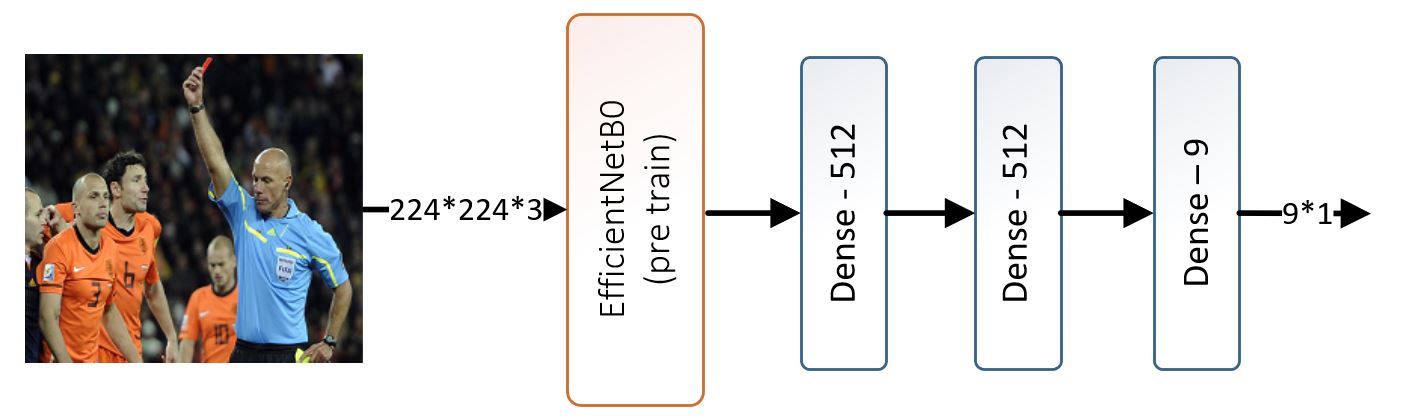}

\caption{Image classification module architecture}
\label{fig:3}

\end{figure*}

\subsection{No Highlight Detection Module (Variational Autoencoder)}

\paragraph A soccer match can include various scenes that are not necessarily a specific event, such as scenes from a soccer match where the director is showing the faces of the players either on the field or on the bench, or when the players are walking and the moments when the game is stopped. These scenes are not categorized as events of a soccer match.

In general, to be able to separate the images of the defined events from the rest of the images, three actions must be performed to complete each other and help us to detect no highlights. The three actions are :
\begin{enumerate}
\item The use of the VAE network to identify if the input images are similar to the \textbf{s}occer \textbf{ev}ent (SEV) dataset images
\item The use of three additional categories, that is, left penalty area, right penalty area, and center circle in the image classification module, given that most free kick images are similar to images from these categories (if these categories were not placed, the images of the wingers of the field would usually get a good score in the free-kick category)
\item Applying the best threshold on the prediction value of the last layer of the EfficientNetB0 feature extraction network.
\end{enumerate}

The second and third methods are applied to the image classification module and will be described later. In the first action, the VAE architecture is employed according to Fig \ref{fig:2} to identify images that do not fall into any of the event categories. To this end, the whole images of the soccer training dataset are given to the VAE network to be trained, then using reconstruction loss and determining a threshold on it, it is determined that the images whose reconstruction loss value are higher than a fixed threshold are not considered as the soccer images. Images with a reconstruction loss value less than a fixed threshold are categorized as the soccer game images and then they are given to the image classification module for classification. In other words, this VAE plays the role of a two-class classifier that puts soccer images in one category and non-soccer images in another category. Reconstruction loss is obtained from the difference between the input image and the reconstructed image. The more input images from a more distinctive distribution, the higher reconstruction loss value will be, and if it is trained using the same image distribution, the amount of this error will be less.

\subsection{Image Classification Module}

The EfficientNetB0 architecture which is shown in Fig \ref{fig:3}  is used to categorize images. This network is responsible for classifying images between nine classes. To place an image in one of the classes of this network, its prediction value at the end layer should be higher than the threshold value which is set 0.9. Otherwise it would be selected as no highlight frame.

If one of the options left penalty area, right penalty area, and center circle is the output of this network, that image is no longer defined as an image of the event that occurred in a soccer match but is included in the no highlight category. However, if it is in one of the categories of penalty kick, corner kick, tackle, free kick, and to substitute, the event will be finalized and decision is made. Finally, if the image is classified in the card category, the image is given to the fine-grain classification module where the color of that card will be determined.

\begin{figure*}[h!]
\centering

\includegraphics[scale=0.6]{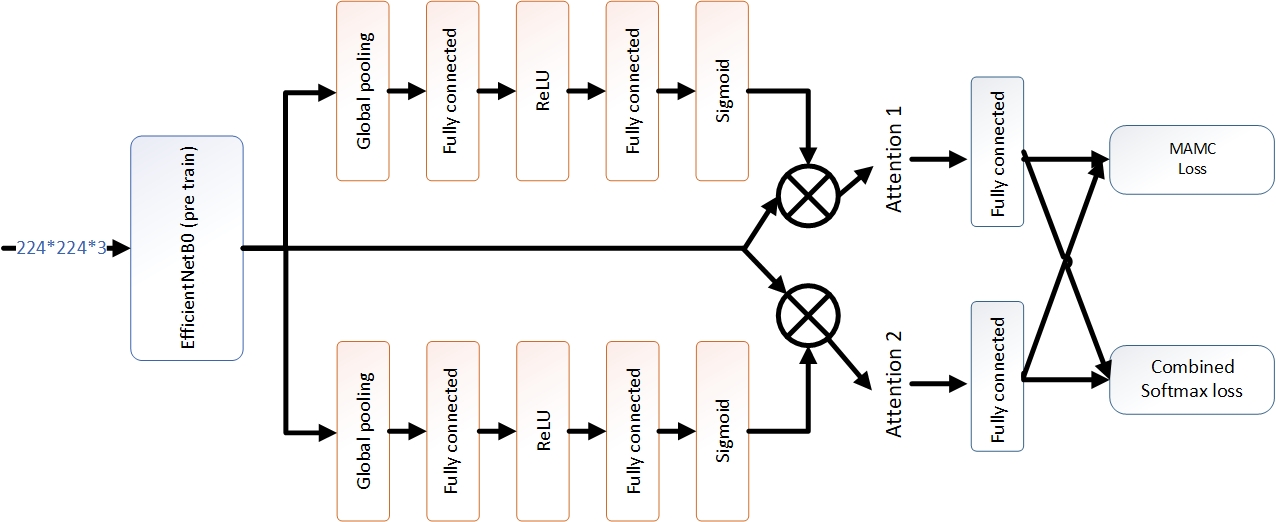}

\caption{Fine-grain classification module architecture}
\label{fig:4}

\end{figure*}

\subsection{Fine-grain Classification Module } 

The only difference between the red and yellow cards is their color of the card; otherwise, there exist no other differences in their image. Thus, both are in the card category but are separate in terms of subcategories. The main classification does not distinguish these two categories well, hence it is decided in the training phase of image classification module to merge these two cards into one category. Also, to separate the yellow and red cards, a separate subclassification is used that focuses on the details. The final architecture employed in this section can be seen in Fig \ref{fig:4} . Here, the proposed architecture provided in \cite{sun2018multi} is exploited, except that instead of the Res-Net50 architecture that is used in \cite{sun2018multi} , the EfficientNetB0 architecture is employed and the network is trained using the yellow and red cards data. The input images to this network are the images that were categorized as cards in the image classification module. The output of this network determines whether these images are in the yellow or red card category.

\begin{figure*}[h!]
\centering

\includegraphics[scale=0.6]{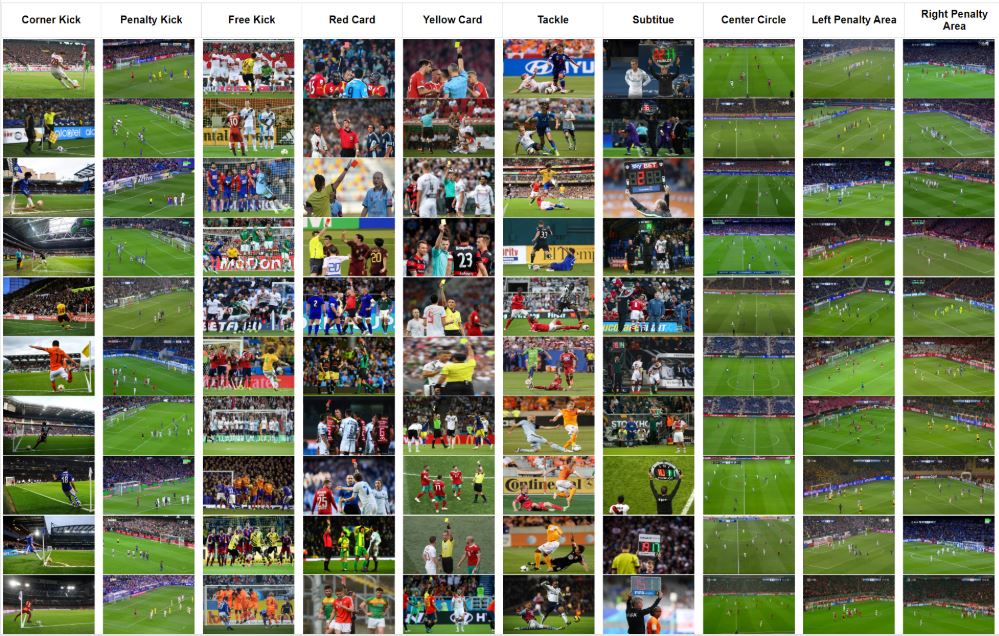}

\caption{Samples of SEV dataset}
\label{fig:5}

\end{figure*}

\section{Datasets}

In this paper, two datasets, that is, soccer event (SEV) and test event datasets have been collected. The collection of these two datasets has been done in two ways: 
\begin{enumerate}
\item By crawling on Internet websites, images of different games were collected. These images are unique and are not consecutive frames
\item Using the video of the soccer games of the last few years of the prestigious European leagues and extracting images related to events.

\end{enumerate}

\subsection{ImageNet}

ImageNet dataset given in \cite{krizhevsky2012imagenet} is employed in the method proposed in this paper for the initial weighting of the EfficientNetB0 network.

\begin{table*}[h!]
\caption{Statistics of SEV dataset}\label{tbl1}
    \makebox[\linewidth]{
\begin{tabular}{lccc}
\toprule
Class Name & \# of train images & \# of validation images & \# of test images\\
\midrule
Corner kick & 5000 & 500 & 500\\
Free kick & 5000 & 500 & 500\\
To Substitute & 5000 & 500 & 500\\
Tackle & 5000 & 500 & 500\\
Red card & 5000 & 500 & 500\\
Yellow card & 5000 & 500 & 500\\
Penalty kick & 5000 & 500 & 500\\
Center Circle  & 5000 & 500 & 500\\
Left Penalty Area & 5000 & 500 & 500\\
Right Penalty Area & 5000 & 500 & 500\\
Sum & 50000 & 5000 & 5000\\

\bottomrule
\end{tabular}
}
\end{table*}

\begin{figure*}[h!]
\centering

\includegraphics[scale=0.5]{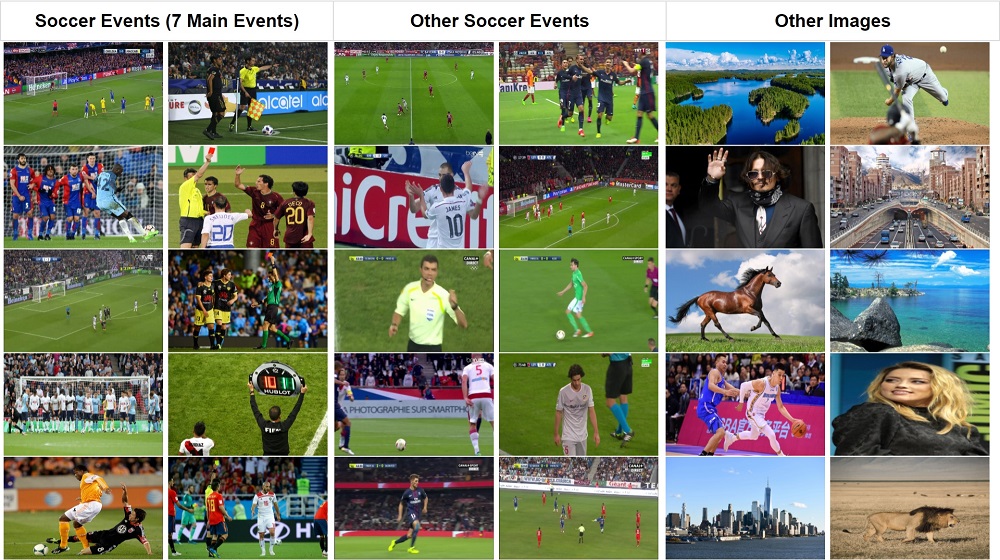}
\caption{Samples of test event dataset}
\label{fig:6}

\end{figure*}

\begin{table*}[h!]
    
    \caption{Statistics of test event dataset}
    \makebox[\linewidth]{
      \begin{tabular}{lc}
        \hline
        Class name & Image instance   \\
         \hline 
        Soccer events (events defined) & 1400	\\
        Other soccer events (throw-in, goal kick, offside, …) & 1400\\
        Other images (images that is not related to football)  & 1400\\
        Sum & 4200\\ 
        \hline
      \end{tabular}
    }
  \end{table*}

\subsection{SEV Dataset}

This dataset includes images of soccer match events that have been collected specifically for this study. In the SEV dataset, a total of 60,000 images were collected in 10 categories. The images of this dataset, as described, were collected in two different ways. Seven of the ten image categories are related to the soccer events defined in this paper, and the rest of the categories are used for no-highlight detection so that the images related to them are not mistakenly included in the seven main categories. Table I shows how dataset SEV is divided into train, validation, and test datasets. Also samples of SEV dataset shown in Fig \ref{fig:5}

\subsection{Test Event Dataset} 

The test dataset is exploited to evaluate the proposed method. Samples of that dataset shown in Fig \ref{fig:6}. The dataset consists of three classes, the first class contains images of the events selected from the SEV dataset, and an equal number of images (200 instances from each category) are selected from each category (only defined events). The second class includes other images of soccer, in which none of these seven events are included. The third class includes images that are not generally related to soccer. Details of the number of images in this dataset are given in Table II.

\section{Experiment and Performance Evaluation}

\subsection{Training}

All three networks are trained independently and the end-to-end method is not employed. The training methods of the 10-class classifier network, the yellow and red card classifier network, and the VAE network are explained in the following subsections, respectively.

\subsubsection{VAE}

\begin{table*}[h!]
    
    \caption{Simulation parameters of the variational autoencoder}
    \makebox[\linewidth]{
      \begin{tabular}{ll}
        \hline
        Parameter & Value   \\
         \hline  \hline
        Optimizer & Adam\\
        Loss function & Reconstruction loss + KL loss\\
        Preformance metric & Loss\\
        \hline
      \end{tabular}
    }
    \label{tab:d1}
\end{table*}

To train this network, the images of seven events defined from the SEV dataset are selected and given to the VAE network as training data. Test and validation data of these seven categories are also used to evaluate the network. The specifications of the simulation parameters used in this network are summarized in Table III.

As illustrated in Fig \ref{fig:7} , the value of the loss curve decreases during successive epochs for validation data.

\subsubsection{Image Classification (9 Class)}{}

\begin{table*}[h!]
    
    \caption{Simulation parameters of the image classification module}
    \makebox[\linewidth]{
      \begin{tabular}{ll}
        \hline
        Parameter & Value   \\
         \hline  \hline
        Optimizer & Adam\\
        Loss function & Categorical Cross-Entropy\\
        Preformance metric & Accuracy\\
        Total Classes & 9 (red and yellow card classes  merged)\\
        Augemnation & Scale, Rotate, Shift, Flip\\
        Batch Size & 16\\ 
        \hline
      \end{tabular}
    }
    \label{tab:d1}
\end{table*}

\begin{table*}[h!]
    
    \caption{Simulation parameters of the fine-grain image classification module}
    \makebox[\linewidth]{
      \begin{tabular}{ll}
        \hline
        Parameter & Value   \\
         \hline  \hline
        Optimizer & Adam\\
        Loss function & Categorical cross-entropy\\
        Preformance metric & Accuracy\\
        Total Classes & 2 (red and yellow card classes)\\
        Augemnation & Scale, rotate, shift, flip\\
        Batch Size & 16\\ 
        \hline
      \end{tabular}
    }
    \label{tab:d1}
\end{table*}

The image classification network is first trained on the ImageNet image collection with dimensions of 224 * 224 * 3. Then, using transfer learning, the network is re-trained on the SEV dataset and is fine-tuned. Dimensions of input images of the SEV dataset are 224 * 224 * 3. The network is trained in 20 epochs with the simulation parameters specified in Table IV. The yellow and red card classes are merged and their 5,000 images are used for network training together with images of other SEV dataset classes.

\subsubsection{Fine-grain image classification}{}

The network shown in Fig \ref{fig:4} is trained using two classes of red card and yellow card of the SEV dataset. Data from each category is partitioned to train, test and validation with 5000, 500, and 500 images. The specifications of the simulation parameters used in this network are given in Table V.

\subsection{Evaluation Metrics}

Different metrics are exploited to evaluate this network. In order to evaluate and compare different image classification architectures of EfficientNetB0 and Fine-grain module networks, the accuracy metric is used as the main metric; recall and F1-score are also used to determine the appropriate threshold value of the EfficientNetB0 network. Also, precision is used to evaluate the performance of the proposed method to detect events.

The accuracy metric can be used to determine how accurately the trained model predicts and, as described in this paper, to compare different architectures and hyperparameters in the EfficientNetB0 and Fine-grain module networks.

\begin{equation}
Accuracy =    \frac{TP  + TN}{P + N} =  \frac{TP + TN}{TP + TN + FP + FN} 
\end{equation}

The F1 score metric considers both the recall and precision criteria together; the value of this criterion is one at the best-case scenario and zero at the worst-case scenario.

\begin{equation}
F_{1}  =   \frac{2}{ Precision^{-1} + Recall^{-1} }  = \frac{TP}{TP +  \frac{1}{2}  \times (FP+FN)} 
\end{equation}

\begin{figure*}[h!]
\begin{subfigure}[b]{0.5\textwidth}
  \includegraphics[width=\linewidth]{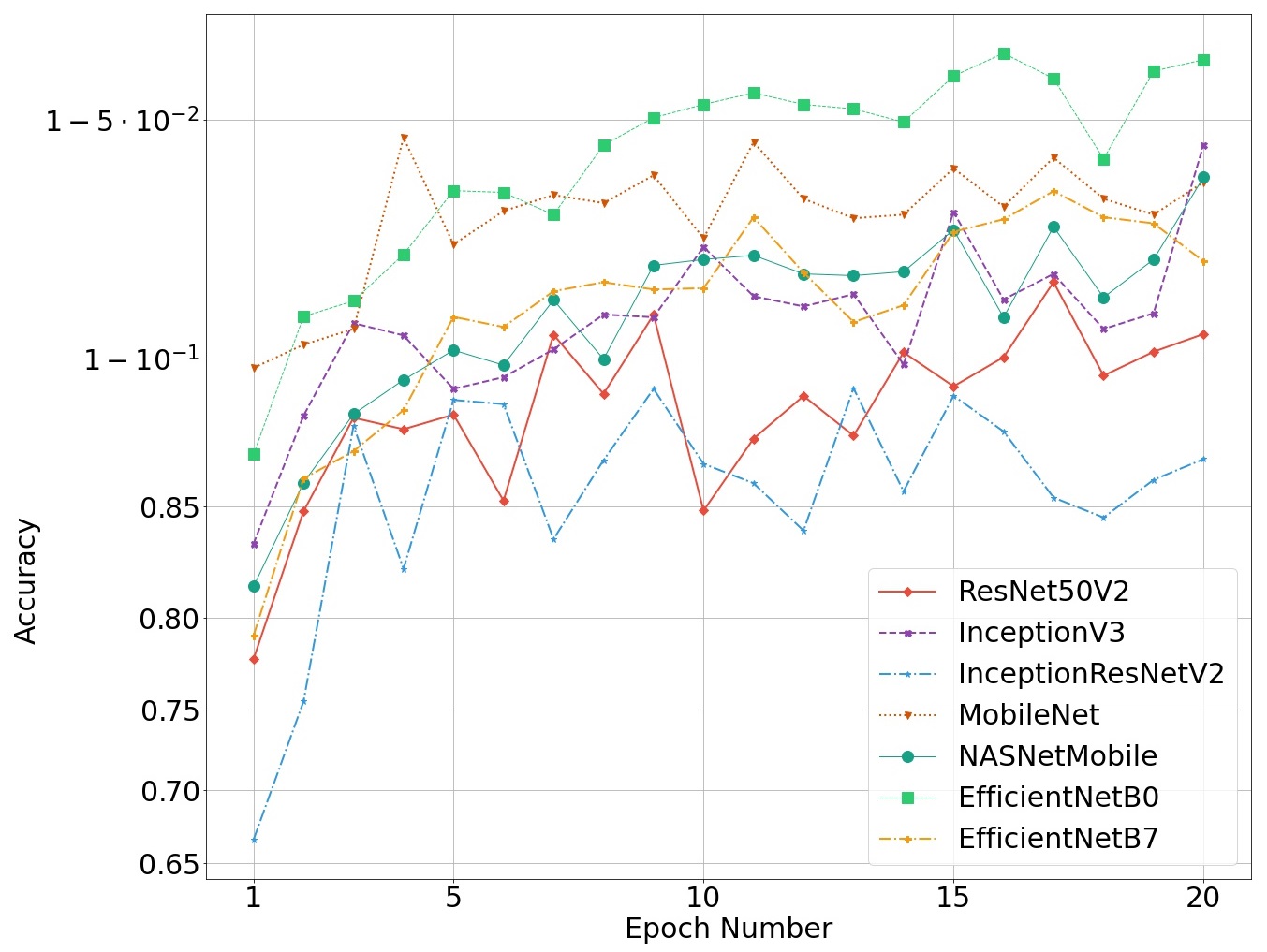}
  \caption{Architecture}
  \label{fig:71}
\end{subfigure}\hfil 
\hfill
\begin{subfigure}[b]{0.5\textwidth}
  \includegraphics[width=\linewidth]{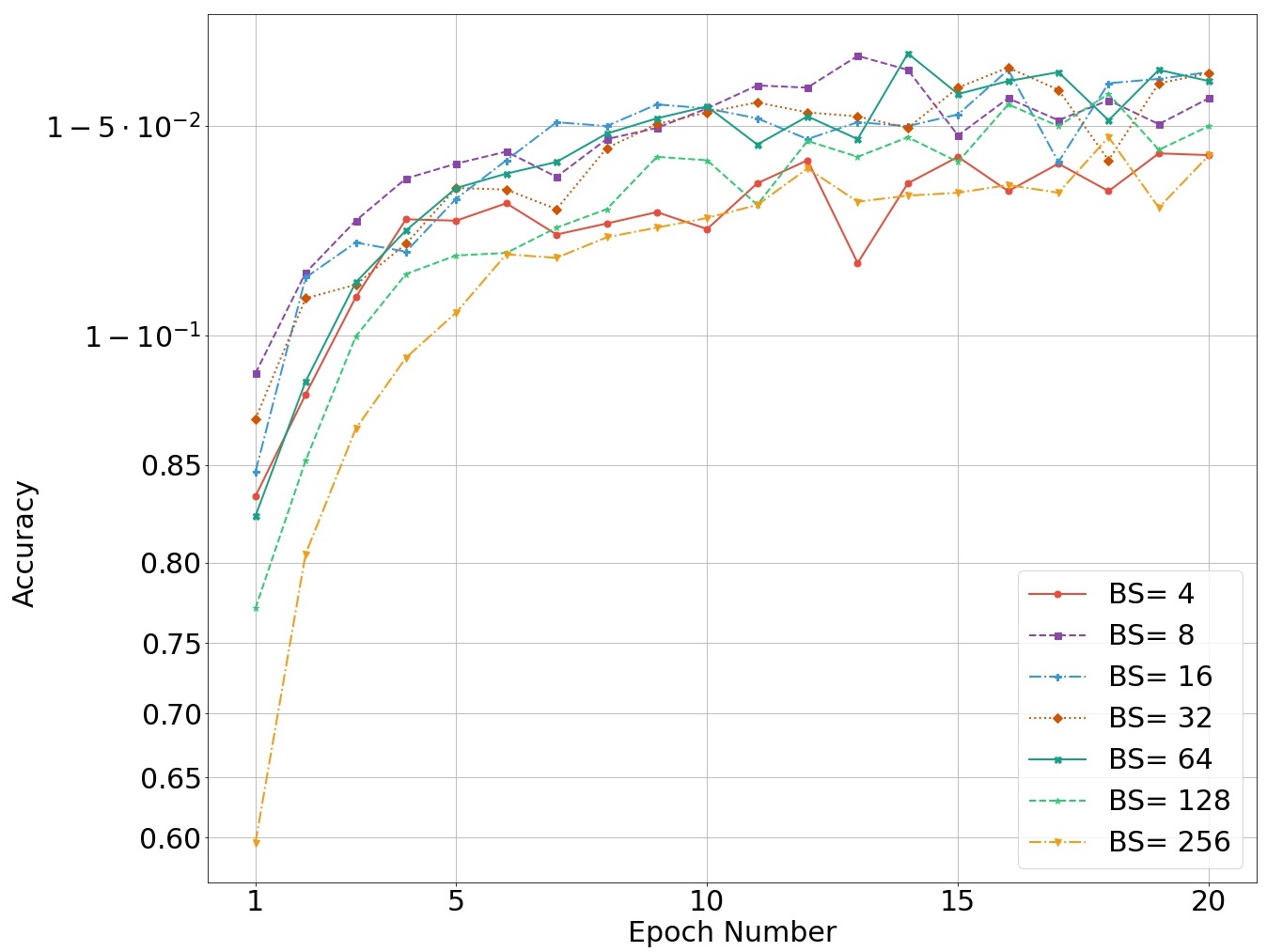}
  \caption{Batch size}
  \label{fig:72}
\end{subfigure}\hfil 
\begin{subfigure}{0.5\textwidth}
  \includegraphics[width=\linewidth]{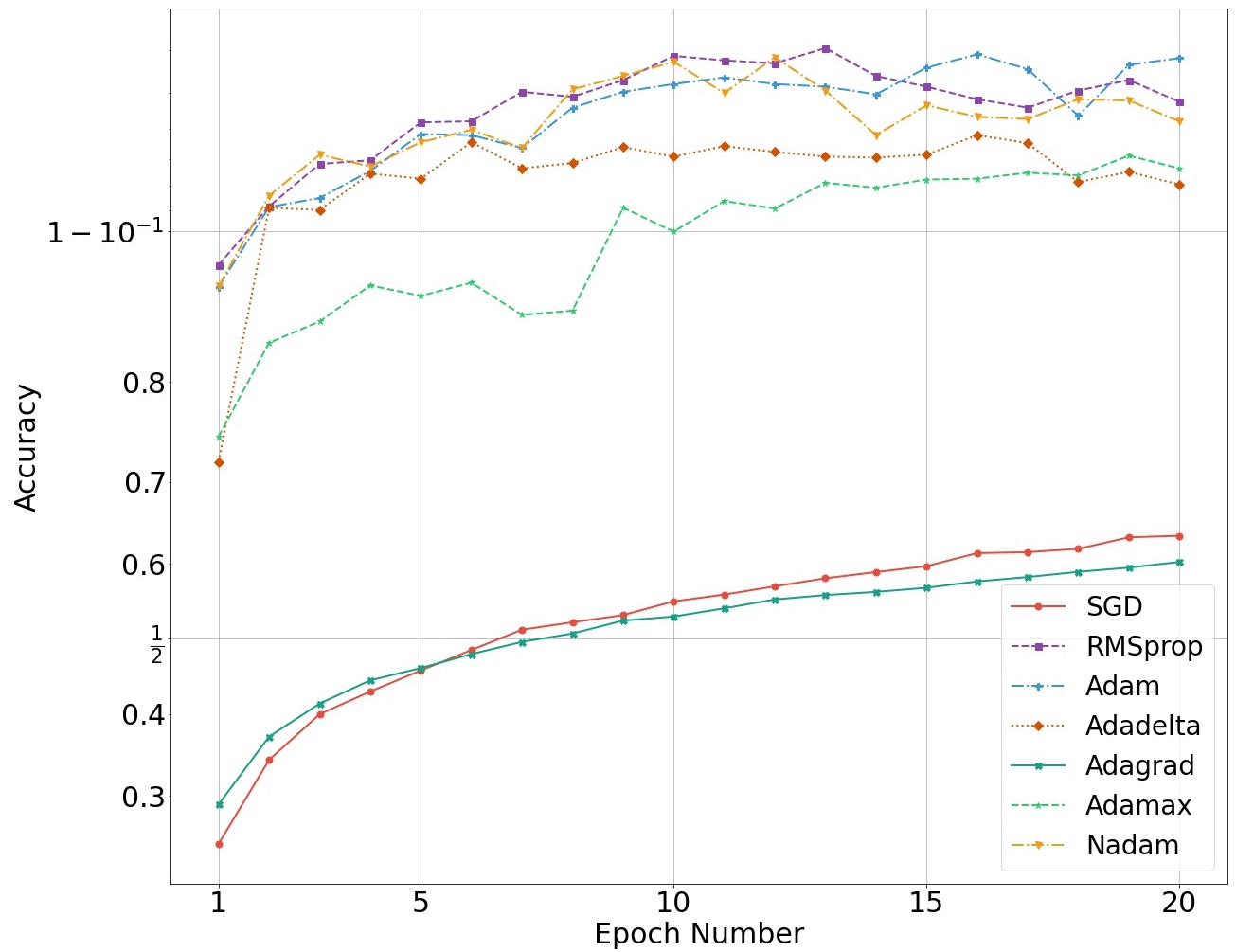}
  \caption{Optimizer}
  \label{fig:73}
\end{subfigure}\hfil 
\begin{subfigure}{0.5\textwidth}
  \includegraphics[width=\linewidth]{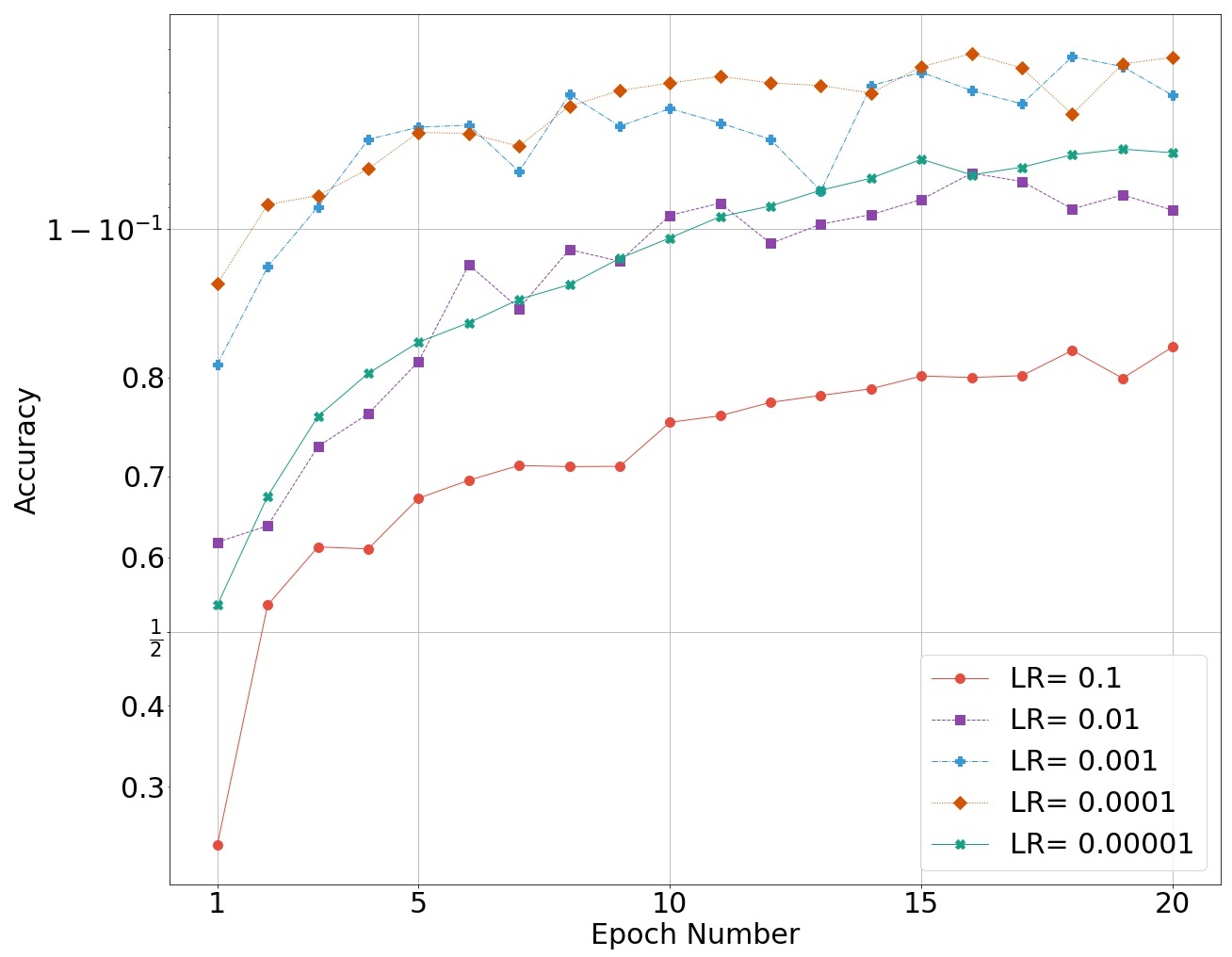}
  \caption{Learning rate}
  \label{fig:74}
\end{subfigure}\hfil 

\caption{Comparing the results on hyperparameters and architecture for validation accuracy (image classification module)}
\label{fig:7}
\end{figure*}

Precision is a metric that helps to determine how accurate the model is when making a prediction. This metric has been used as a criterion in selecting the appropriate threshold.

\begin{equation}
Precision =  \frac{TP}{TP + FP} 
\end{equation}

Recall metric refers to the percentage of total predictions that are correctly categorized .

\begin{equation}
Recall =  \frac{TP}{TP + FN} 
\end{equation}

\subsection{Evaluation}

The various parts of the proposed method have been evaluated to achieve the best model in order to detect an event in a soccer match. In the first step, the algorithm should be able to effectively classify the images of the defined events correctly. In the next step, the network is examined to see how the network can detect no highlights, and the best possible model is selected. Eventually, the performance of the proposed algorithm in a
soccer video is examined and compared to other state-of-the-art methods.

\begin{table*}[h!]
    
    \caption{Comparison between 
    Fine-grain image classification models}
    \makebox[\linewidth]{
      \begin{tabular}{lccc}
        \hline
        Method name &	Epoch &	Acc (red and yellow card) (\%) &	Acc (CUB-200-2011) (\%)  \\
         \hline  \hline

        B-CNN \cite{lin2015bilinear}\cite{tan2019efficientnet} & 60 & 66.86 & 84.01  \\        
        OSME + MAMC using Res-Net50 \cite{sun2018multi} & 60 & 61.70 & 86.2 \\
        \textbf{OSME + MAMC \cite{sun2018multi} using EfficientNetB0} \cite{tan2019efficientnet} & 60 & \textbf{79.90} & 88.51  \\
        Ge et. al. \cite{ge2019weakly} & 60 & 78.03 & \textbf{90.40} \\
        \hline
      \end{tabular}
    }
    \label{tab:d1}
\end{table*}

\subsubsection{Classification evaluation}{}

The image classification module is responsible for classifying images. To test and evaluate this network,  as shown in Fig \ref{fig:7}, different architectures were used for training and different hyperparameters were also tested for these models. As shown in Fig \ref{fig:71} , the EfficinetNetB0 model has the best accuracy among the other models.

If in the above model, we divide the card images into two categories of yellow and red cards and give the dataset to the network in the form of the same 10 categories of SEV datasets for training, the accuracy of test data is reduced from 94.08\% to 88.93\% . The reason for this is the interference of yellow and red card predictions in this model. Consequently, the yellow and red cards detection have been assigned to a subclassification, and only the card category in the EfficientNetB0 model has been used.

For the subclassification of yellow and red card images, various fine-grained methods were evaluated and the results are shown in  Table VI. The bilinear CNN method \cite{lin2015bilinear} using the EfficientNet architecture achieves 66.86\% accuracy and the OSME method that employs the EfficientNet architecture reaches 79.90\% accuracy, which shows higher accuracy than the other methods. Nonetheless, the main architecture (EfficientNetB0) used in the image classifier shows 62.02\% accuracy, which gives difference of 17.88\%. 

Table VII compares the accuracy of combining the image classification module and fine-grain classification module for image classification with other models. As shown in Table VII, the accuracy of our proposed method for image classification is 93.21\%, which shows the best accuracy among all models. Also, the proposed method is still faster than the other models except MobileNet and MobileNetV2. To prove this point, The execution time of each model is calculated for 1400 images, and their average run time as the mean inference time is given in Table VII.

\begin{table*}[h!]
    
    \caption{Comparison between image classification models on SEV dataset}
    \makebox[\linewidth]{
      \begin{tabular}{cccc}
        \hline
        Method name & Accuracy (\%) &	Mean inference time (second)\\
         \hline  \hline

        VGG16 \cite{simonyan2014very} & 83.21 & 0.510 \\
        ResNet50V2 \cite{he2016identity}  & 84.18 & 0.121 \\
        ResNet50 \cite{he2016deep}  & 84.89 & 0.135 \\
        InceptionV3 \cite{szegedy2016rethinking} & 84.93 & 0.124 \\
        MobileNetV2 \cite{sandler2018mobilenetv2} & 86.95 & 0\textbf{.046} \\    
        Xception \cite{chollet2017xception}  & 86.97 & 0.186 \\
        NASNetMobile \cite{zoph2018learning} & 87.01 & 0.121 \\
        MobileNet \cite{howard2017mobilenets} & 88.48 & 0.048 \\
        InceptionResNetV2 \cite{szegedy2016inception} & 88.71 & 0.252 \\      
        EfficientNetB7  \cite{tan2019efficientnet}  & 88.92 & 1.293 \\
        EfficientNetB0  \cite{tan2019efficientnet}  & 88.93 & 0.064 \\
        DenseNet121 \cite{huang2017densely} & 89.47 & 0.152 \\

       \textbf{Proposed method (image classification section)}  & \textbf{93.21} & 0.120 \\
        \hline
      \end{tabular}
    }
    \label{tab:d1}
\end{table*}

As shown in Table VIII, the problem of card overlap is also solved, and the proposed method in the image classification section separates the two categories almost well.

\begin{table*}[h!]
    
    \caption{Confusion matrix proposed algorithm (image classification section)}
    \makebox[\linewidth]{
      \begin{tabular}{ccccccccccc}
        \hline
  &  &  &  & Predicted &  &  &  &  &  & \\
Actual & Center &	Corner &	Free kicke &	Left &	Penalty &	R Card &	Right &	Tackle &	To Substitue &	Y Card \\
        \hline  \hline

        Center & 0.994 & 0 & 0.006 & 0 & 0 & 0 & 0 & 0 & 0 & 0 \\
        Corner & 0 & 0.988 & 0.006 & 0 & 0 & 0.004 & 0 & 0 & 0.002 & 0 \\
        Free kick & 0.004 & 0.004 & 0.898 & 0.03 & 0 & 	0.002 & 0.05 & 0.006 & 0.004 & 0.002 \\

        Left Penalty Area & 0.002 & 0 & 0.036 & 0.958 & 0.004 & 0 & 0 & 0 & 0 & 0 \\

        Penalty kick & 0 & 0 & 0 & 0.002 & 0.972 & 0 & 0.026 &	0 & 0 & 0  \\

        Red cards & 0 & 0.008 & 0.03 & 0 & 0 & 0.862 & 0 & 0.002 & 0.012 & 0.086 \\

        Right Penalty Area & 0 & 0 & 0.01 & 0.002 & 0 & 0 & 0.988 & 0 & 0 & 0 \\

        Tackle & 0.012 & 0.014 & 0.02 & 0.02 & 0.002 & 0.006 & 0.028 & 0.876 & 0.004 & 0.018 \\

        To Substitue & 0 & 0 & 0 & 0 & 0 & 0.002 & 0.002 & 0 & 0.994 & 0.002  \\

        Yellow cards & 0 & 0.022 & 0.01 & 0 & 0 & 0.131 & 0 & 0.002 & 0.006 & 0.829 \\

        \hline
      \end{tabular}
    }
    \label{tab:d1}
\end{table*}

\begin{table*}[h!]
    
    \caption{Comparison between different threshold values (image classification module)}
    \makebox[\linewidth]{
      \begin{tabular}{ccc}
        \hline
        Threshold &	F1-score (event images) (\%) & Recall (images not related to football) \%) \\
        \hline  \hline

        0.99  & 86.6 & 95.3 \\
        0.98  & 88.2 & 94.2   \\
        
       0.97  & 89.2 & 93.8  \\
        
       0.95  & 90.5 & 93.1  \\

        0.90  & 91.8  & 92.4 \\
       0.80  & 92.4 & 76.9  \\

       0.50  & 95.2  & 51.2  \\

        \hline
      \end{tabular}
    }
    \label{tab:d1}
\end{table*}

\subsubsection{Known or Unknown Evaluation}

In order to determine the threshold value on the output of the last layer of the EfficientNetB0 network, according to Table IX, different threshold values were tested and evaluated; and then the value 0.90 which gives the highest sum of F1-score for detecting the event and also gives the best recall to detect no highlight images was determined as the threshold for the last layer of the network.

\begin{figure*}[h!]
\centering

\includegraphics[scale=0.3]{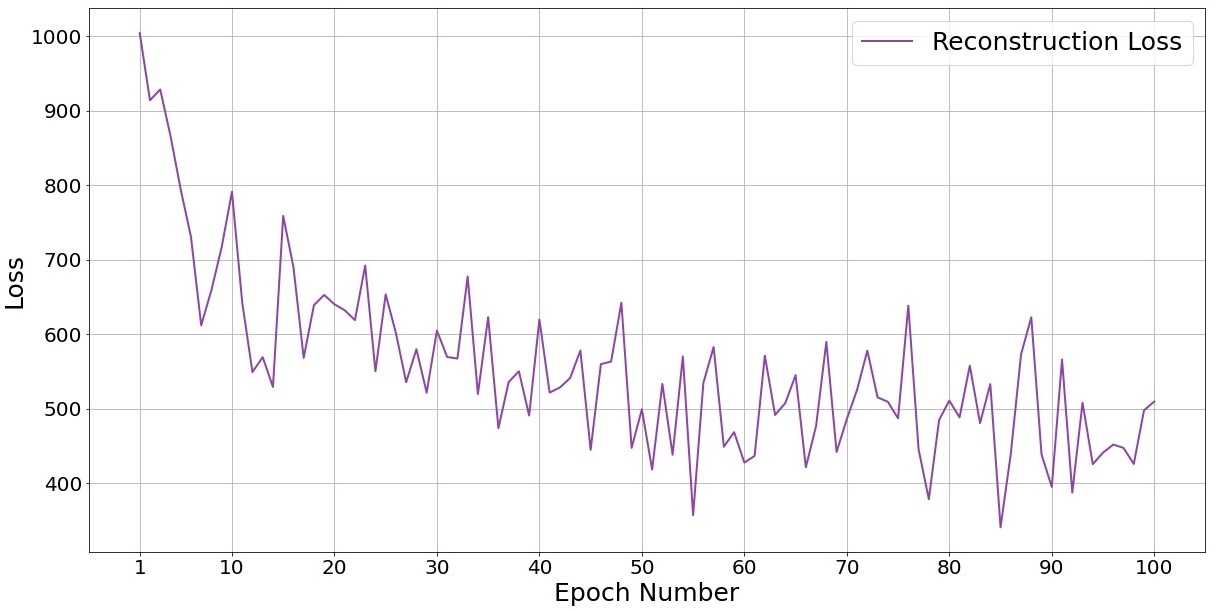}

\caption{Reconstruction Loss of VAE}
\label{fig:8}

\end{figure*}

The threshold value for the loss of VAE network was also examined with different values as shown in Fig \ref{fig:8} , the value of 328 as the threshold for the loss gives the best distinction between categories.

To test how the network detects no highlight images and the defined event images, the test dataset is used. Using this dataset helps to know the number of main events incorrectly classified by the proposed method as events not related to soccer. Also, it clarifies the number of images related to soccer classified as the main events, and the number of images categorized in the no highlight category. Moreover, it specifies the number of images related to the real event categorized correctly into the correct events, and the number of events that are not included in the main events. The results of this evaluation are provided in Table X.

\begin{table*}[h!]
    
    \caption{The precision of the proposed algorithm}
    \makebox[\linewidth]{
      \begin{tabular}{ccc}
        \hline
        Class & Sub-class & Precision\\

        \hline  \hline

       Soccer Events & Corner kick & 0.94 \\
        --- & Free kick & 0.92 \\
        ---  & To Substitue & 0.98 \\
        ---  & Tackle & 0.91 \\
        ---  & Red Card & 0.90 \\
        ---  & Yellow Card & 0.91 \\
        ---  & Penalty Kick & 0.93 \\
        Other soccer events &  & 0.86 \\
        Other images &  & 0.94 \\

        \hline
      \end{tabular}
    }
    \label{tab:d1}
\end{table*}

\subsubsection{Final Evaluation}

Ten soccer matches have been downloaded from the UEFA Champions league and then, using the proposed method, the task of events detection has been carried out. In this evaluation, events that occur in each soccer game are examined. 
In other words, the number of events correctly detected, and the number of events incorrectly detected by the network have been determined. Details of the results are given in Table XI and compared with other similar methods. in state-of-the-art methods

\begin{table*}[h!]
    
    \caption{Precision of the proposed and state-of-the-art methods on 10 soccer matches (\%)}
    \makebox[\linewidth]{
      \begin{tabular}{cccc}
        \hline
        Event name & Proposed method & BN\cite{tavassolipour2013event} & Jiang et. al.\cite{jiang2016automatic}\\

        \hline  \hline

        Corner kick & 94.16 & 88.13 & 93.91\\
        Free kick & 83.31 & \--- & \--- \\
        To Substitue & 97.68 & \--- & \--- \\
        Tackle & 90.13 & 81.19 & \--- \\
        Red Card & 92.39 & \--- & \--- \\
        Yellow Card & 92.66 & \--- & \--- \\
        Card & \--- & 88.83 &  93.21 \\
        Penalty Kick &  88.21 & \--- & \--- \\

        \hline
      \end{tabular}
    }
    \label{tab:d1}
\end{table*}

\section{Conclusions and Future Work}

In this paper, two  novel datasets for soccer event detection has been presented. One is the SEV dataset including 60,000 images in 10 categories, seven of which were related to soccer events, and three to soccer scenes; these were used in training image classification networks. The images of this dataset were taken from top fine leagues in Europe and the European Champions league. The other dataset is the test event that contains 4200 images. Test event includes three categories: the first category consists of the events mentioned in the paper, the second category comprises other images of a soccer match apart from the first category events, and the third category includes images off the soccer field. This dataset was exploited to examine the network power in detecting and distinguishing between images with highlight and those with no highlight.

Furthermore, a method for soccer event detection is proposed. The proposed method employed the EfficientNetB0 network in the image classification module to detect events in a soccer match. Also, the fine-grain image classification module was used to differentiate between red and yellow cards. If this module is not employed, red and yellow cards would have been categorized in the image classification module, and the differentiation accuracy would have been 88.93\%. However, the fine-grain image classification module increased the accuracy up to 93.21\%. In order to solve the network problems in predicting the images other than those defined, a VAE was employed to adjust the value of the threshold and several images, other than those of the defined events, were used to make a better distinction between the images of the defined events and other images.

\bibliography{mybibfile}
\bibliographystyle{IEEEtran}

\begin{IEEEbiography}[{\includegraphics[width=1in,height=1.25in,clip,keepaspectratio]{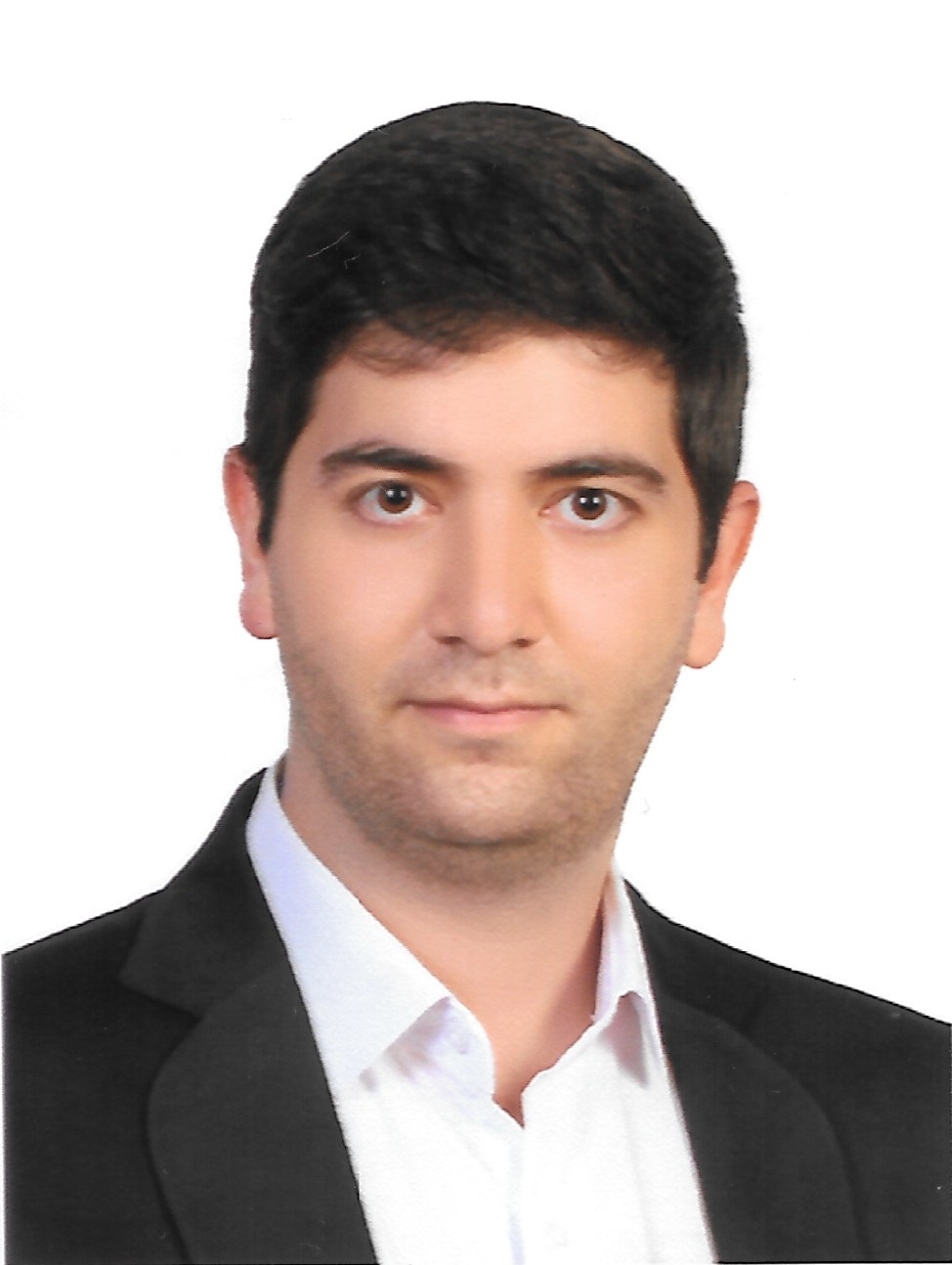}}]{Ali Karimi}
received the B.S. degree in Computer Engineering (software engineering) from Bu-Ali Sina University, Hamedan, Iran, in 2018. He is currently pursuing the M.S. degree in Information Technology Engineering at University of Tehran, Tehran, Iran.
His fields of interest include Image \& Video Processing, Machine Vision and Machine Learning.
\end{IEEEbiography}
\begin{IEEEbiography}[{\includegraphics[width=1in,height=1.25in,clip,keepaspectratio]{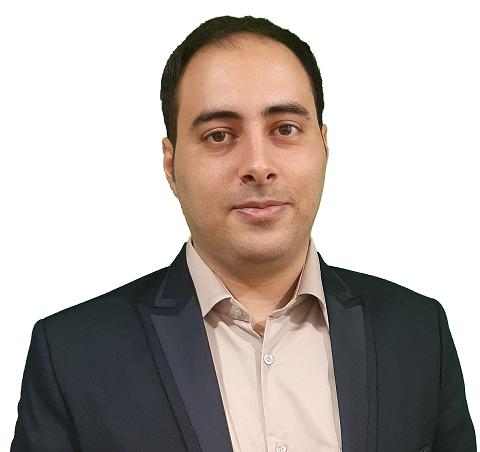}}]{Ramin Toosi}
received the B.Sc. degree in Electrical Engineering (communication devision) from Shahid Beheshti University, Tehran, Iran in 2014. He received the M.Sc. degree in Electrical Engineering (system division) from University of Tehran in 2016. He is currently pursuing the Ph.D. degree in Communications Systems at the Secure Communication Lab, University of Tehran. His fields of interest include Signal Processing, Multimedia Security, Pattern Recognition, and Information Theory.
\end{IEEEbiography}
\begin{IEEEbiography}[{\includegraphics[width=1in,height=1.25in,clip,keepaspectratio]{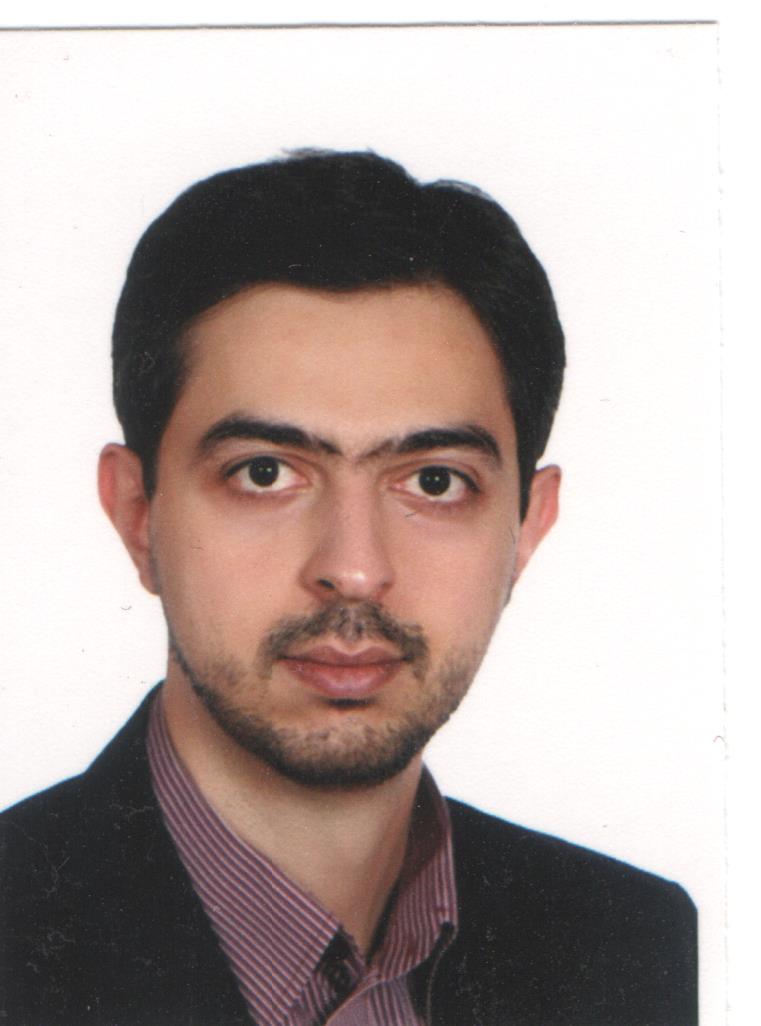}}]{Mohammad Ali Akhaee}
(S’07–M’07) received the B.Sc. degree in Electronics and
Communications Eng. from the Amirkabir University of Technology, Tehran, Iran, and
the M.Sc. and Ph.D. degrees from the Sharif University of Technology, Tehran, Iran, in
2005 and 2009, respectively. He is currently an Assistant Professor with the College of
Engineering and the Director of the Secure Communication Laboratory, University of
Tehran, Tehran, Iran. He has authored or coauthored more than 60 papers, and holds
one Iranian patent. His research interests include the area of signal processing, in
particular multimedia security, data hiding, and machine learning.
Prof. Akhaee was the Technical Program Chair of EUSIPCO ’11 and the Executive
Chair of ISCISC ’14 and Financial Chair of RTEST’18. He received the Governmental
Endeavour Research Fellowship from Australia in 2010 and the Governmental award
from Ministry of Information and Communication Technology (ITC) from Iran in 2017.
\end{IEEEbiography}

\end{document}